# ISWSST: Index-space-wave State Superposition Transformers for Multispectral Remotely Sensed Imagery Semantic Segmentation

Chang Li [a], Pengfei Zhang [a] *, Yu Wang [a]

[a] *Key Laboratory for Geographical Process Analysis & Simulation of Hubei Province, and*

*College of Urban and Environmental Science, Central China Normal University, Wuhan, China*

*\* Corresponding author: Pengfei Zhang, zhangpf@mails.ccnu.edu.cn*

**Abstract**

*Currently the semantic segmentation task of multispectral remotely sensed imagery (MSRSI) faces the following problems: 1) Usually, only single domain feature (i.e., space domain or frequency domain) is considered; 2) downsampling operation in encoder generally leads to the accuracy loss of edge extraction; 3) multichannel features of MSRSI are not fully considered; and 4) prior knowledge of remote sensing is not fully utilized. To solve the aforementioned issues, an index-space-wave state superposition Transformer (ISWSST) is the first to be proposed for MSRSI semantic segmentation by the inspiration from quantum mechanics, whose superiority is as follows: 1) index, space and wave states are superposed or fused to simulate quantum superposition by adaptively voting decision (i.e., ensemble learning idea) for being a stronger classifier and improving the segmentation accuracy; 2) a lossless wavelet pyramid encoder-decoder module is designed to losslessly reconstruct image and simulate quantum entanglement based on wavelet transform and inverse wavelet transform for avoiding the **edge extraction loss**; 3) combining multispectral features (i.e. remote sensing index and channel attention mechanism) is proposed to accurately extract ground objects from original resolution images; and 4) quantum mechanics are introduced to interpret the underlying superiority of ISWSST. Experiments show that ISWSST is validated and superior to the state-of-the-art architectures for the MSRSI segmentation task, which improves the segmentation and edge extraction accuracy effectively. Codes will be available publicly after our paper is accepted.*

## 1. Introduction

Recently, Transformer architectures have achieved excellent performance in different remote sensing vision tasks, such as image classification, object detection and semantic segmentation [1]. The heart of Transformer architectures is multi-head attention mechanism, which translates 2D image-based tasks into 1D sequence-based tasks. Benefiting by the powerful capability of sequence-to-sequence (long-range) relational modelling, Transformer architecture demonstrates superior performance of extracting global context compared to convolutional neural networks (CNNs), achieving state-of-the-art results on the fundamental semantic segmentation task [2].

In current research, various domain features are extracted with remarkable Transformer architectures to performance high-accuracy semantic segmentation: (1) spatial domain feature is extracted with Transformer, e.g., VIT [3], Swin-Transformer [4], SegFormer [42], Unet-Transformer [6], and PVT [7], among others; (2) frequency domain feature is extracted with Transformer using Wave-MLP [8], Wave-ViT [9], and other methods. In addition, more multichannel features of multispectral remote sensing image (MSRSI) are further mined by the following approaches: (1) many remote sensing indexes are designed based on the spectral characteristics of ground objects, which can extract ground objects from original resolution images (e.g., NDVI [7], NDBI [10], NDISI [11], NDWI [8], NDBSI [12], etc.); (2) channel attention mechanism is designed to modeling the multi-channel relationship effectively, which is rarely used in multispectral images. Based on the above research status, there are some issues for MSRSI semantic segmentation which still need to be further addressed: (1) only space domain feature or frequency domain feature is usually considered; (2) edge extraction loss is generated by the downsampling operations; (3) the mining of multichannel features of MSRSI is insufficient; (4) prior knowledge of remote sensing index is not combined with deep learning; and (5) deep learning usually is regarded as a "black box" lacking reasonable interpretation.

Quantum mechanics has been being widely recognized as the greatest innovation of the 21st century and has promoted a giant leap forwards especially in the fields of quantum communication, quantum sensing, and quantum computation [13]. It descripts the physical properties of nature and express the real-world essence at the microscopic scale. The effects of quantum mechanics include wave–particle duality, quantum entanglement, and quantum superposition, which inspire us new insights into solving the aforementioned issues about semantic segmentation as below: (1) the pixel of an image in spatial domain and frequency spectrum of an image in wave domain could be regarded as particle and wave respectively corresponding to wave–particle duality; (2) quantum entanglement could be simulated by the wavelet transform and inverse wavelet transform, e.g., transform between spatial and frequency domains; (3) quantum superposition simulation could be introduced by inspiring from ensemble learning to make decision fusing spatial, frequency and index domain features; and (4) quantum mechanics could further interpret the underlying rationality of our proposed network. Therefore quantum mechanics has the great potential to solve the current bottleneck of semantic segmentation.

Based on above principles of quantum mechanics, we propose an index-space-wave state superposition Transformer (ISWSST) for MSRSI semantic segmentation, which is the first quantum-inspired multi-superposition-state fusion net framework. The contributions and advantages of this proposed ISWSST are as follows:

(1) Multiple states of (i.e., index, space and wave domains) are the first to be superposed or fused to simulate quantum superposition by adaptive voting decision liking ensemble learning for MSRSI semantic segmentation. Exactly, index-space-wave state superposition could complement the advantages of others to take superiority of similar ensemble learning in deep learning. Meanwhile, an adaptive voting decision/ensemble learning module is designed to dynamically optimize the decision weights. It is conducive to scientifically make decision and fuse feature of each domains for improving the generalization and segmentation accuracy.

(2) A lossless wavelet pyramid encoder-decoder (LWPED) module is the first to be designed to extract lossless edges and simulate quantum entanglement by wavelet transform and inverse wavelet transform between spatial and frequency domains. The wavelet transform and inverse wavelet transform are utilized to replace the downsampling, which are embedded in encoder to reconstruct loss images. That improves the accuracy edge extraction effectively.

(3) Combining multispectral features (i.e., remote sensing index and channel attention mechanism) is the first to be proposed. Multiple remote sensing indexes extracted from the original resolution images are introduced to as prior knowledge to compensate edge errors. And channel attention mechanism could automatically optimize multispectral information and reduce the channel noise to improve the perception ability for the channel details. That are more helpful to improve the segmentation and edge extraction accuracy.

(4) The interpretable ISWSST is further to be thoroughly revealed under the guidance of multiple quantum effects from the two aspects: the overall accuracy improvement and edge accuracy improvement. It's a significant exploration to descript the scientificity and rationality of ISWSST combining quantum machines simulation.

## 2. Related Work

**Multispectral-based semantic segmentation.** MSRSI covers a wider part of the electromagnetic spectrum, which is beneficial to recognize ground objects. CNNs have brought a revolution in modeling multispectral information and extracting more effective abstract visual features such as FCN[14], U-Net[15] and DeepLabV3[16]. It is worth noting that more and more research is focusing on the channel attention mechanism to more effectively model long-term dependencies of multi-channel information, such as HCRB-MSAN [17], CCAFFMNet [18], DSPCANet [19], CAM-DFCN [20], RELAXNet [21] and SE-Net) [22]. And it represents the valuable potential to improve accuracy in semantic segmentation.

**Remote-sensing-index-based semantic segmentation.** By utilizing the band selection and algebraic operations for MSRSI, remote sensing index can be constructed for rapid extraction of specific category information of land use/land cover (LULC), such as Normalized Difference Vegetation Index (NDVI) [7], Normalized Difference Built-up Index (NDBI) [10], Normalized Difference Impervious Surface Index (NDISI) [11], Normalized Difference Water Index (NDWI) [8] and Normalized Difference Bare Soil Index (NDBSI) [12], among others. Based on these above remote sensing indexes, LULC category information of ground objects can be quickly identified without prior knowledge. However, the classification accuracy of the above index-based methods is still limited by the influence of the phenomenon that the same objects have different spectral features and different objects have same spectral features. Besides, current researches lack the effective integration of the semantic segmentation methods based on remote sensing indexes and deep learning.

**Transformer-based semantic segmentation architectures.** In recent years, Transformer structures have demonstrated excellent performance in semantic segmentation tasks compared to CNNs [9] [23] [24] [3] [25]. Vision Transformer (ViT) [3] is originally proposed for image recognition. Originally proposed for image recognition, Vision Transformer (ViT) [3] focuses more on extracting global information compared to the ability of CNNs to extract local details. The multi-headed attention mechanism integrated within Transformers prevents overfitting, making them independent of each other and capable of obtaining richer feature information, resulting in better semantic segmentation results. Multi-head attention allows the model to jointly attend to information from different representation subspaces at different positions. Transformers such as Longformer [26], Transformer iN Transformer (TNT) [27], HCTransformers[28] and NomMer [29] demonstrate the potential in current MSRSI semantic segmentation tasks, with further possibilities for exploration.

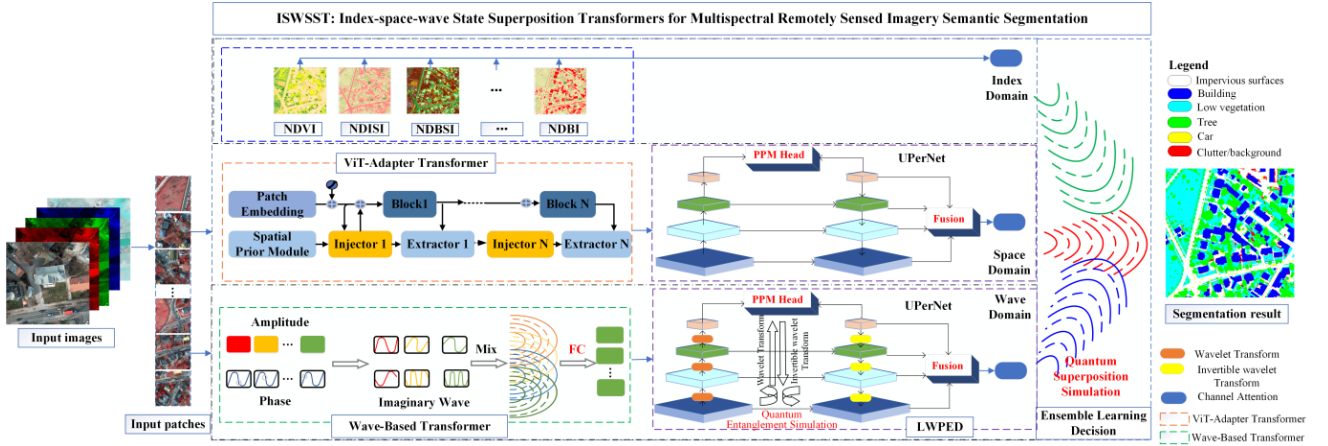

Figure 1: Overall architecture of this proposed ISWSST. Firstly, index, space and wave domain features are extracted; secondly, a lossless wavelet pyramid encoder-decoder (LWPED) module is embedded in encoder-decoder of wave domain to reconstruct lossless multi-scale features; pyramid pooling module (PPM) is introduced in the UPerNet; finally, ensemble learning decision is designed to superposed multiple domain features and obtain high-accuracy result.

**Wave-based Transformer semantic segmentation Architectures.** In previous studies on Transformer, researchers began to try to use frequency domain information with Wave-based Transformer for visual tasks. And the architectures have achieved good results. Tang [30] proved that information in image processing can be computed by Wave-based Transformer. Tang [32] achieved efficient semantic segmentation by introducing wavelet transform considering the balance of efficiency and accuracy at same time. In summary, Wave-based Transformers have proven to be an efficient scheme for semantic segmentation using frequency information, and have obtained competitive results in a variety of dense prediction tasks. Therefore, fully exploiting and using frequency information have significant potential for improving semantic segmentation tasks.

**Quantum machine learning**. Recently, some scholars have begun to focus on quantum machine and introduced it to perform dense prediction tasks. Tang [32] introduced a wave function with amplitude and phase as an quantum entity to dynamically modulating the relationship between tokens and fixed weights in MLP. It indicates that Quantum mechanics is a rational and prominent approach to resolve the semantic segmentation problems. Nonetheless, there exist several urgent bottlenecks ignored in the current research which could be further contemplated combining with quantum effects of quantum machine. For example, quantum superposition could be introduced to model the relationship among the multiple domain feature decision, quantum entanglement could be introduced to compensate the loss of edge extraction by the lossless transform between spatial domain and frequency domain and quantum mechanics is introduced to enhanced interpretability of underlying mechanism and behaviors of our proposed network from a micro point of view.

## 3. Methodology

### 3.1. The overall of the workflow

In the section, the overall workflow of this proposed ISWSST network is described detailly in Figure 1, which includes the flowing steps:

(1) Index-space-wave domain feature extraction is performed to simulate multiple quantum states, which aims to yield output features for predicting the pixel-level probability of segmented objects.

(2) A LWPED module is designed to simulate quantum entanglement for transforming spatial domain information to be frequency domain information, which replace the downsampling operation in encoder with the wavelet transform and inverse wavelet transform to implement lossless multiple-scale feature transform.

(3) Remote sensing index and channel attention mechanism are designed to mine multispectral features for reducing edge extraction loss and improving segmentation accuracy. Multiple remote sensing indexes are extracted from original resolution images, which means that the objects could maintain the lossless edge information. And the channel attention mechanism is embedded in after output features of every domain for enhancing the relationship between the various channel features.

(4) A multiple state voting decision/ensemble learning module is designed to simulate quantum superposition for yielding segmentation result. Above obtained output

·3

features of index-space-wave domains are adaptively superposed or fused with dynamic voting weights liking ensemble learning for optimizing the effect of multiple domain integration.

The main design details are shown as the following.

### 3.2. Index-space-wave domain feature extraction

Space and wave domain features are constructed based on the encoder-decoder architecture. UPerNet [33] is selected as the decoder to obtain the multiple scale fused features based on the pyramid pooling module (PPM).

**Index domain construction.** A series of remote sensing indexes are designed based on multispectral information of MSRSIs, including NDVI [7], NDBI [10], NDISI [11], NDWI [8] and NDBSI [12], etc.

**Space domain construction.** ViT-Adapter [1] network is selected as the encoder to extract various scale semantic features, which yielded state-of-the-art performance in many semantic segmentation tasks. It contains the following modules: (1) the original architecture of ViT [3] is adopted to extract the plain multi-scales visual deep features; (2) spatial prior module (SPM) is designed to enhance the local spatial contexts; (3) spatial feature injector (SFI) is proposed to dynamically adjust the relationship of the features of SPM and the ViT [3]. (4) Multi-scale feature extractors utilize the cross-attention layer and the feed-forward network (FFN) to yield multi-scale features. The detailed diagram is shown in ViT-Adapter Transformer module of Figure 1.

**Wave domain Construction.** Wave-based Transformer is designed as the encoder to obtain various scale semantic features in wave domain which could transform the spatial information from an image to be frequency information. For the Transformer architecture, an input image generally is divided into image patches as $X = [x_1, x_2, \cdots, x_n]$, where each patch $x_j$ is a d-dimension vector. They are converted into waves (i.e., $\tilde{x}_j$) with the amplitude and phase information, i.e.:

$$\tilde{x}_j = |x_j| \odot e^{-i\theta_j}, j = 1,2\ldots, n \quad (1)$$

where $i$ represents the imaginary unit, which satisfies this condition $i^2 = -1$. $|\cdot|$ represents the absolute value operation. $\odot$ is means that element-wise multiplication operation. $|x_j|$ represents the amplitude of each patch, $\theta_j$ represents the phase information, which is regarded as the current location of the image patch within a wave period. With the Euler's formula, Eq.1 can be unfolded and the wave-like image can be represented in the complex domain with real part and imaginary part, i.e.,

$$\tilde{x}_j = |x_j| \odot \cos\theta_j + i\,|x_j| \odot \sin\theta_j, j = 1,2\cdots, n \quad (2)$$

The wave-based deep visual feature extraction relies on Wave-based Transformer module shown in Figure 1, including channel-mixing MLP and token-mixing MLP. In order to combine information from different channels, channel-mixing MLP yield the communications along the channel dimension and is formulated based on ChannelFC function as below:

$$x_j = \text{ChannelFC}(x_j, W^c) = W^c x_j, j = 1,2\cdots, n \quad (3)$$

where $W^c$ represents the weight with learnable parameters. Combing information from different tokens, token-mixing MLP layers obtain the communications along the spatial dimension. Then, global information can be obtained by fusing the information of multiple patches using Fourier series, which is the real-value output $u_j$ by summing the real and imaginary part of $x_j$ with weights.

$$u_j = \sum_k \left(W^t_{jk} x_k \odot \cos\theta_k + W^i_{jk} x_k \odot \sin\theta_k\right),$$
$$j = 1,2 \ldots, n \quad (4)$$

where $W^t$, $W^i$ are both learnable weights. In the above equation, the phase $\theta_k$ is dynamically optimized based on the semantic content of input data. Besides the fixed weights, the phase also modulates the aggregating process of different patches (i.e., tokens).

### 3.3. A LWPED module

The wavelet transform and inverse wavelet transform are designed to alleviate the edge accuracy loss due to the scale transformation (i.e., downsampling operation) in the encoding stage.

In wave domain, Wavelet transform is embedded in encoding stage, and inverse wavelet transform is adopted in decoding stage. Haar wavelet [34] is selected as the Discrete Wavelet Transform function. A discrete image is assumed as $\{s_j(k_1, k_2)\}$ and the low-pass filter $h$ and the high-pass filter $g$ are respectively preformed for each row of $s_j$, and $k_1$ and $k_2$ are the width and height of the image. And they are performed for each column of $s_j$. Then the low-frequency component $s_{j+1}$ and the high-frequency components $d^h_{j+1}, d^v_{j+1}, d^d_{j+1}$ are obtained. Then wavelet transform [34] is defined as the following:

$$s_{j+1}(n_1, n_2) = \sum_{k_1}\sum_{k_2} \left(h(2n_1 - k_1)h(2n_2 - k_2)s_k(k_1 - k_2)\right) \quad (5)$$

$$d^h_{j+1}(n_1, n_2) = \sum_{k_1}\sum_{k_2} \left(h(2n_1 - k_1)g(2n_2 - k_2)s_k(k_1 - k_2)\right) \quad (6)$$

$$d^v_{j+1}(n_1, n_2) = \sum_{k_1}\sum_{k_2} \left(g(2n_1 - k_1)h(2n_2 - k_2)s_k(k_1 - k_2)\right) \quad (7)$$

$$d^d_{j+1}(n_1, n_2) = \sum_{k_1}\sum_{k_2} \left(g(2n_1 - k_1)g(2n_2 - k_2)s_k(k_1 - k_2)\right) \quad (8)$$

where $n_1$ and $n_2$ are the width and height of the decomposed image components.

The inverse wavelet transform is defined as the following in Eq.8, where $\tilde{h}$ and $\tilde{g}$ are respectively represented as the corresponding coefficient of high-pass filter and low-pass filter.

The wavelet transform and inverse wavelet transform can be represented respectively as the formulate of matrix, i.e., Ep.9 and Ep.10, where $A$ is the input feature vector, $B$ is the output feature vector and $H$ is the transformation matrix.

$$s_j(k_1, k_2) =$$
$$\sum_{n_1}\sum_{n_2}\left(\tilde{h}(k_1-2n_1)\tilde{h}(k_2-2n_2)s_{j+1k}(n_1-n_2)\right) +$$
$$\sum_{n_1}\sum_{n_2}\left(\tilde{h}(k_1-2n_1)\tilde{g}(k_2-2n_2)d_{j+1}^h(n_1,n_2)\right) +$$
$$\sum_{n_1}\sum_{n_2}\left(\tilde{g}(k_1-2n_1)\tilde{h}(k_2-2n_2)d_{j+1}^v(n_1,n_2)\right) +$$
$$\sum_{n_1}\sum_{n_2}\left(\tilde{g}(k_1-2n_1)\tilde{g}(k_2-2n_2)d_{j+1}^d(n_1,n_2)\right) \quad (9)$$
$$B = HAH^T \quad (10)$$
$$A = H^TBH \quad (11)$$

### 3.4. Channel attention guided feature selection

Channel attention mechanism [35] is performed to select the key information and generate the weights of channel attention for the output feature of index, special and wave domains. Then the aforementioned out feature maps could be regarded as the input feature maps and be optimized. The statistical expressions of channel attention are as follows:

$$E_{avg} = (1/W \times H)\sum_{i=1}^{W}\sum_{j=1}^{H} L_s(i,j) \quad (12)$$

where $L_s$ represents the input feature map with spatial dimensions $W \times H$, $L_s(i,j)$ represents the value of the input feature map with the location $(i,j)$. $E_{avg}$ indicates the channel-wise mean of the feature map having dimensionality as $E_{avg} \in \mathbb{R}^{1 \times 1 \times C}$. The sigmoid activation ($\sigma$) over mean channel features ($E_{avg}$) produces channel-attentive weights, which further enhances spatial features to yield channel-attentive feature maps $O_C \in \mathbb{R}^{F \times d_{model}}$ as:

$$\emptyset_c = L_s \odot \sigma(W_c(dr(E_{avg}))) \quad (13)$$

where $W_c$ indicates the learnable weight for channel-wise attention operation. Finally, channel-attentive feature maps $\emptyset_c$ are flatten to represent 2D data as $O_c \in \mathbb{R}^{F \times d_{model}}$.

### 3.5. Multiple state voting decision/ensemble learning

The above output features of three domains selected by channel attention mechanism can be regarded as heterogeneous quantum state features, where $S$, $F$, and $I_n$ are the output features of space, wave, and index domains respectively and $I_n$ represents that $n$th index of remote sensing indexes. To simulate quantum state superposition, multiple quantum state features can be superimposed together to complement each other's strengths from different domains. Finally, a voting decision approach liking ensemble learning with adaptive weights can be constructed:

$$\varphi = \text{argmax}(\lambda_1\text{Softmax}(S) + \lambda_2\text{Softmax}(F)$$
$$+\lambda_3\text{Softmax}(I_1) + \cdots + \lambda_{n+2}\text{Softmax}(I_n)) \quad (19)$$

where $\varphi$ is the segmentation result. $\lambda_1 \ldots \lambda_{n+2}$ are the dynamic weights for adaptively adjusting the results of the multi-domain output features.

## 4. Experiments

In the section, the effectiveness of this proposed ISWSST architecture is evaluated in the MSRSI semantic segmentation task, through a series of empirical experiments. Concretely, various deep learning state-of-the-art (SOTA) architectures for semantic segmentation (i.e., vision Transformers, MLPs and CNNs) are compared with this proposed ISWSST architecture to quantify these learnt visual representations in excellent MSRSI dataset (i.e., ISPRS Potsdam and Vaihingen dataset). Finally, ablation studies are performed to verify the contribution of each principal component.

### 4.1. Dataset

In the task of remote sensing semantic segmentation, classical and outstanding multispectral benchmark datasets are selected, which have been widely used to test many of the best deep learning networks.

**ISPRS Potsdam 2D Dataset.** The dataset is an outstanding multispectral airborne image dataset provided by the International Society for Photogrammetry and Remote Sensing (ISPRS), which has four multispectral bands (i.e., near infrared, red, green and blue). It includes 38 fine spatial resolution true ortho photo (TOP) titles, whose ground sampling distance is 9 cm. And the dataset covers six types of LULC, namely, impervious surfaces (IS), buildings (B), low vegetation (LV), trees (T), cars (C), and clutter/background (C). The dataset is divided into training dataset with 23 titles, validation dataset with 1 title and testing dataset with 14 titles. The size of each title is 2494×2064 pixels, and each title was cropped into the fixed size images (i.e., 512×512 pixels) using a sliding window striding 256 pixels. Because the adopted dataset only has four spectral bands, NDVI can be yielded as the RS index.

**ISPRS Vaihingen 2D Dataset.** The dataset is a SOTA multispectral airborne image dataset provided by ISPRS, which has three multispectral bands (i.e., red, green, blue, and near infrared). The dataset is grouped into the same six categories as ISPRS Potsdam 2D Dataset. It includes 33 true ortho photo (TOP) titles, whose ground sampling distance is 9 cm. The dataset is divided into training dataset with 17 titles, validation dataset with 1 title and testing dataset with 15 titles. The size of each title is 2494×2064 pixels, and each title was cropped into the fixed size images (i.e., 512×512 pixels) using a sliding window striding 256 pixels. Because the adopted dataset only has three spectral bands, NDVI can be yielded as the RS index.

## 4.2. Setting

In the MSRSI semantic segmentation task, the experiments are conducted on the ISPRS Potsdam and Vaihingen dataset. All compared SOTA architectures and this proposed ISWSST architecture are trained and optimized on the training and validation dataset, and accuracy metrics are evaluated on the testing dataset. The following metrics are shown to represents the performance of the above architectures, namely, over accuracy (OA), mean intersection over union (mIoU). In the training stage, the pre-trained weights of ImageNet are used, the AdamW optimizer was selected to optimize the networks. Besides, the networks are fine-tuned for 80k iterations with the same batchsize. And the polynomial learning rate scheduler was conducted with the initial learning rate 0.001. The dataset is trained on 4 NVIDIA 3090Ti GPUs.

## 4.3. Results

The evaluation results of different networks (i.e., compared SOTA networks and this proposed ISWSST network) for MSRSI semantic segmentation task are detailly represented in the following Table 1. It's worth noting that the all evaluation metrics of this proposed architecture outperforms the current two mainstream SOTA networks (i.e., CNNs and Transformers) consistently on the two MSRSI datasets. Compared with the best Transformer architectures in the table, this proposed network represents a large superiority, which obtains 88.9% mIoU and 93.8% OA, 85.9% mIoU and 93.0 % OA respectively on ISPRS Potsdam and Vaihingen 2D Dataset. Similarly, compared with the best CNN architectures in the table, this proposed network also represents a greater superiority. The above results concretely demonstrate the superiority of ISWSST architecture for MSRSI semantic segmentation task.

Table 1: Comparison of this proposed ISWSST architecture with existing vision semantic segmentation models on ISPRS Potsdam 2D Dataset.

| Method | Backbone | OA | mIoU |
|---|---|---|---|
| PSPNet [36] | ResNet-101[2] | 90.2 | 84.4 |
| DeepLabV3+[38] | ResNet-101[2] | 91.8 | 85.9 |
| HRNet [39] | HRNet-W48 [39] | 90.7 | 84.9 |
| OCRNet [40] | HRNet-W48 [39] | 93.0 | 87.6 |
| MaskFormer [41] | Swin-Small [4] | 93.3 | 87.5 |
| Swin [4] | Swin-Base [4] | 93.1 | 87.2 |
| Segmenter [5] | ViT-Large [3] | 92.5 | 86.5 |
| SegFormer [42] | MiT-B4 [42] | 92.8 | 86.9 |
| SegFormer [42] | CSWin [43] | 93.2 | 88.1 |
| UPerNet [33] | ViT-Adapter [23] | 93.4 | 88.5 |
| UPerNet [33] | Wave-MLP [8] | 93.3 | 88.3 |
| Ours proposed ISWSST | | 93.8 | 88.9 |

Table 2: Comparison of this proposed ISWSST architecture with existing vision semantic segmentation models on ISPRS Vaihingen 2D Dataset.

| Method | Backbone | OA | mIoU |
|---|---|---|---|
| PSPNet [36] | ResNet-101[2] | 89.8 | 80.4 |
| DeepLabV3+[38] | ResNet-101[2] | 91.1 | 83.2 |
| HRNet [39] | HRNet-W48 [39] | 90.2 | 82.9 |
| OCRNet [40] | HRNet-W48 [39] | 91.9 | 83.8 |
| MaskFormer [41] | Swin-Small [4] | 92.3 | 84.5 |
| Swin [4] | Swin-Base [4] | 92.1 | 84.2 |
| Segmenter [5] | ViT-Large [3] | 91.7 | 84.5 |
| SegFormer [42] | MiT-B4 [42] | 92.0 | 84.9 |
| SegFormer [42] | CSWin [43] | 92.2 | 85.1 |
| UPerNet [33] | ViT-Adapter [23] | 92.4 | 85.2 |
| UPerNet [33] | Wave-MLP [8] | 92.2 | 85.0 |
| Ours proposed ISWSST | | 93.0 | 85.9 |

## 4.4. Ablation Studies

**The formulation of index-space-wave state encoding module.** The integrated different domain features can complement each other's advantages and reduces the impact of noise in the meantime. Different architectures are designed to investigate the effectiveness in Table 3. Especially, the remote sensing index domain obviously guides to enhance the segmentation results. Without the integrated different domain features, the performance of only single networks is obviously inferior, with the best 88.5% mIoU and 93.4% OA on ISPRS Potsdam 2D Dataset, and the best 85.2% mIoU and 92.4% OA on ISPRS Vaihingen 2D Dataset. This proposed index-space-wave state encoding module yields much better performance (e.g., 88.9% mIoU and 93.8% OA on ISPRS Potsdam 2D Dataset, and 85.9% mIoU and 93.0% OA on ISPRS Vaihingen 2D Dataset).

Table 3: Comparison of the effectiveness of index-space-wave state encoding module with existing vision semantic segmentation models on ISPRS 2D Dataset.

| Method | Backbone | Dataset | OA | mIoU |
|---|---|---|---|---|
| UPerNet [33] | ViT-Adapter [23] | Potsdam | 93.4 | 88.5 |
| UPerNet [33] | Wave-MLP [8] | Potsdam | 93.3 | 88.3 |
| Ours proposed ISWSST | | Potsdam | 93.8 | 88.9 |
| UPerNet [33] | ViT-Adapter [23] | Vaihingen | 92.4 | 85.2 |
| UPerNet [33] | Wave-MLP [8] | Vaihingen | 92.2 | 85.0 |
| Ours proposed ISWSST | | Vaihingen | 93.0 | 85.9 |

**The formulation of the LWPED module.** The LWPED module is a flexible module which can be embedded in any computer vision tasks to instead of the upsamping/downsampling operation. Four simple formulations are designed to estimate the performance. The results are shown in Table 4, which shows that the architectures equipped inverse wave-based block achieve much better mIoU values. In summary, this proposed module is effective for reducing **edge extraction loss** and can be widely applied for semantic segmentation tasks to improve the accuracy.

Table 4: Comparison of the effectiveness of LWPED module with existing vision semantic segmentation models on ISPRS Potsdam 2D Dataset.

| Method | Dataset | OA | mIoU |
|---|---|---|---|
| ISWSST with inverse wave-based block | Potsdam | 93.8 | 88.9 |
| ISWSST without inverse wave-based block | Potsdam | 93.5 | 88.4 |
| ViT-Adapter [23]+ UPerNet [33] with inverse wave-based block | Potsdam | 93.7 | 88.6 |
| ViT-Adapter [23]+ UPerNet [33] without inverse wave-based block. | Potsdam | 93.4 | 88.5 |

**The formulation of state-superposition block.** To effectively coordinate the relationship among the discrepant feature domains, the state-superposition block is implemented. It is compared with the classical ensemble learning strategies (i.e., majority voting [17], average voting [18]) shown in Table 5. It's noting that this proposed module aggregates the effect of feature fusion. The segmentation accuracy is higher than other ensemble learning strategies. That's means that this dynamic integration scheme fully takes the relationship among each feature domain into account and effectively improves the segmentation results.

Table 5: Comparison of the effectiveness of state-superposition block on ISPRS Potsdam 2D Dataset.

| Method | Dataset | OA | mIoU |
|---|---|---|---|
| ISWSST with state-superposition block | Potsdam | 93.8 | 88.9 |
| ISWSST with majority voting | Potsdam | 93.2 | 88.3 |
| ISWSST with average voting | Potsdam | 93.0 | 88.1 |

**The formulation of channel attention module.** To further mine the relationship of multi-channel remote sensing multispectral information, the channel attention mechanism is embedded in this proposed ISWSST network. Two simple formulations are designed to show the function of the module, i.e., ISWSST with channel attention; ISWSST without channel attention. It indicates that the channel attention mechanism makes the ISWSST architecture obtains the most significant gains in mIoU and OA values than the model without channel attention mechanism which is shown in Table 6.

Table 6: Comparison of the effectiveness of channel attention module on ISPRS 2D Dataset.

| Method | Dataset | OA | mIoU |
|---|---|---|---|
| ISWSST with channel attention | Potsdam | 93.8 | 88.9 |
| ISWSST without channel attention | Potsdam | 93.6 | 88.5 |
| ISWSST with average voting | Vaihingen | 93.0 | 85.9 |
| ISWSST without channel attention | Vaihingen | 92.6 | 84.2 |

### 4.5. Discussion

The effectiveness and rationality of our ISWSST introducing quantum effects is thoroughly discussed from the two aspects:

(1) The high-accuracy interpretability of edge extraction

The proposed LWPED module in the ISWSST is based on quantum entanglement effect, which realizes the lossless feature transform to replace downsampling operations in encoding stage. On the other hand, remote sensing index maintains the high-accuracy edge details as the original resolution images, which can further compensate the edge errors of the output results of space and wave domains.

(2) The high-accuracy interpretability of OA

The proposed ISWSST leverages ensemble learning approach to simulate quantum superposition for adaptive fusing output features of index, space and wave domains. Space feature focus on the spatial relation of pixels, and wave feature focus on the location relation of frequency spectrums, which are independent and complementary. In addition, the optimization of edge extraction accuracy directly leads to the OA improvement.

## 5. Conclusion

The paper proposes ISWSST for MSRSI semantic segmentation. The multiple domain features (i.e., index domain, space domain and wave domain) are firstly superposed to integrate the advantages of each domain and reduce the noise. In addition to this, a LWPED module is designed to instead of the scale transform operations, which can be embedded in any architecture and can reduce the **edge extraction loss**. It's noting that multispectral feature combination (i.e. remote sensing index and channel attention mechanism) is proposed to further mine the multispectral channel information and improve the edge extraction accuracy. Extensive experiments indicate that this proposed ISWSST is a strong architecture for the MSRSI dense prediction task (i.e., semantic segmentation). In the future, we will further explore the potential of ISWSST architecture on more downstream tasks, such as object detection, instance segmentation.

## Acknowledgements

This work is supported by the National Natural Science Foundation of China (NSFC) under Grant Nos. 41771493 and 41101407, and the Fundamental Research Funds for the Central Universities under Grant CCNU22QN019.